\documentclass[pmlr,twocolumn,10pt]{jmlr}

\usepackage{booktabs}
\usepackage{siunitx}
\usepackage{mathrsfs}
\usepackage{xspace}
\usepackage{enumitem}
\usepackage{float}

\newif\ifcomments

\commentstrue

\ifcomments
    \newcommand{\chenhao}[1]{{\color{blue}{\tt #1 --CT}}}
    \newcommand{\dang}[1]{{\color{red}{\tt #1 --DN}}}
    \newcommand{\hh}[1]{{\color{orange}{\tt #1 --HH}}}
\else
    \newcommand\chenhao[1]{}
    \newcommand\dang[1]{}
    \newcommand\hh[1]{}
\fi

\newcommand{\xray}{X-ray\xspace}
\newcommand{\secref}[1]{\S\ref{#1}\xspace}

\theorembodyfont{\upshape}
\theoremheaderfont{\scshape}
\theorempostheader{:}
\theoremsep{\newline}

\jmlrvolume{225}
\jmlryear{2023}
\jmlrsubmitted{LEAVE UNSET}
\jmlrpublished{LEAVE UNSET}
\jmlrworkshop{Machine Learning for Health (ML4H) 2023} 

 \title[Pragmatic Radiology Report Generation]{Pragmatic Radiology Report Generation}

 \author{
    \Name{Dang Nguyen} \Email{dangnguyen@uchicago.edu} \\
    \addr The University of Chicago
    \AND
    \Name{Chacha Chen} \Email{chacha@uchicago.edu} \\
    \addr The University of Chicago
    \AND
    \Name{He He} \Email{hhe@nyu.edu} \\
    \addr New York University
    \AND
    \Name{Chenhao Tan} \Email{chenhao@uchicago.edu} \\
    \addr The University of Chicago
 }

\begin{document}

\maketitle

\begin{abstract}

When pneumonia is not found on a chest \xray, should the report describe this negative observation or omit it?
We argue that this question cannot be answered from the \xray alone and requires a pragmatic perspective, 
which captures the communicative goal that radiology reports serve between radiologists and patients.
However, the standard image-to-text formulation for radiology report generation fails to incorporate such pragmatic intents. 
Following this pragmatic perspective, we demonstrate that the indication, which describes why a patient comes for an \xray, drives the mentions of negative observations.
We thus introduce indications as additional input to report generation. 
With respect to the output, we develop a framework to identify uninferable information from the image, which could be a source of model hallucinations, and limit them by cleaning groundtruth reports.
Finally, we use indications and cleaned groundtruth reports to develop pragmatic models, and show that they outperform existing methods not only in new pragmatics-inspired metrics (e.g., +4.3 Negative F1) but also in 
standard metrics (e.g., +6.3 Positive F1 and +11.0 BLEU-2).

\end{abstract}

\section{Introduction}
\label{sec:introduction}

\begin{table}[t]
\centering
\small
\begin{tabular}{ p{0.45\textwidth} } 
\toprule
INDICATION:  An \_\_\_-year-old woman with previous aspiration pneumonia and a history of congestive heart failure (CHF). \\
IMPRESSION:  \textcolor{orange}{PA and lateral chest} \textcolor{blue}{compared to \_\_\_:} Lungs are hyperinflated, due to airway obstruction or emphysema. \textcolor{orange}{On the lateral view}, aside from a granuloma, \textcolor{red}{there is no pneumonia. The heart size is normal, no pulmonary edema related to CHF}. Right pleural effusion is tiny \textcolor[HTML]{1589FF}{status post pleural tube removal} \textcolor{blue}{compared to large pleural effusions seen on prior chest radiographs}. \textcolor{red}{There are no findings to suggest intrathoracic malignancy.} \textcolor[HTML]{9400D3}{
An urgent CT thorax is suggested given the rapid growth of granuloma.
} 
\textcolor[HTML]{006400}{These findings were communicated to Dr. \_\_\_ at 4:00 p.m. by phone.} \\
\bottomrule
\end{tabular}
\caption{A synthetic chest \xray report, highlighting that a report includes more than positive findings from \xray.
\textcolor{blue}{Blue}: prior comparisons. \textcolor[HTML]{1589FF}{Light blue}: previous procedures. \textcolor{red}{Red}: negative mentions. \textcolor{orange}{Orange}: image view. \textcolor[HTML]{006400}{Green}: doctor communication.
\textcolor[HTML]{9400D3}{Purple}: medical recommendations. 
}
\label{tab:intro_example}
\end{table} 

Radiology report generation
has emerged as an important problem in machine learning for healthcare~\citep{jing2018automatic, wang2018tienet, liu2019clinically, yuan2019multiview, chen2020generating, endo2021retrieval, miura2021improving, ramesh2022improving, thawkar2023xraygpt, tu2023medpalmm}.
In particular, MIMIC-CXR~\citep{johnson2019mimic} is a widely used dataset due to its large number of \xray images and corresponding radiology reports.

In this work, we revisit the standard formulation of radiology report generation and the MIMIC-CXR benchmark from a pragmatic perspective.
Radiology report generation is typically formulated as an image-to-text problem: generate a complete report given a chest \xray. 
We argue that this formulation does {\bf not} align with the functional goal of radiology reports as a communicative device between medical professionals and patients~\citep{hartung2020create}.

\begin{table*}[t]
\small
\centering
    {\begin{tabular}{p{3cm}p{7cm}p{5.0cm}}
     \toprule
     \bf Input Factor &  \bf Input Description & \bf Relevant Content  \\
    \midrule
    X-ray image(s) & The image(s) taken for the current study & Positive observations \\
    \midrule
    \multicolumn{3}{c}{Factors Beyond the Images}\\
    \midrule
    Indication & The reason for a patient's visit & Positive \& negative observations \\
    Previous studies & Findings from previous chest X-rays & Comparisons to prior studies \\
    Previous treatment, medical history & Medical procedures the patient has received & Mentions of previous procedures \\
    Communication information & Communication between medical professionals, electronic systems & Mentions of what information transfer has taken place \\
    Image view & The X-ray view(s) from which a patient is seen & Mentions of the view, often before commenting on findings \\
    Medical expertise/situation & Medical expertise \& knowledge about the patient's preference/other conditions & Medical recommendations \\
    \bottomrule
    \end{tabular}}
    \caption{Categorization of the types of input that can influence a radiology report. Examples of each type of output can be seen in in Table~\ref{tab:intro_example}.
    }
    \label{tab:input_factors}
\end{table*} 
To illustrate, Table~\ref{tab:intro_example} shows an example report. 
The very first line is {\em Indication}, where the radiologist explains why the patient needed a chest \xray.
This information is not part of the image, but plays a crucial role in determining the content of the report.
One example is mentions of negative diagnosis (henceforth \emph{negative mentions}): although one might infer that any unnamed observation is negative, this is not how radiologists communicate with each other or 
with patients.
In Table~\ref{tab:intro_example}, the sentences with negative mentions highlighted in red (``there is no pneunomia'' and ``no pulmonary edema related to CHF'') specifically respond to the conditions ``pneumonia'' and ``CHF'' in the indication.
In contrast, 
other common conditions such as Pneumothorax are omitted.

In general, radiologists convey much more information than positive findings from an image and the pragmatic perspective is critical to understand what makes a radiology report.
Table~\ref{tab:input_factors} provides a comprehensive view of different factors that may affect a report's content.
Notably, there are many factors beyond the image itself.\footnote{Image views may be learnable from the data with images from different views. However, most studies only use a single image as the input and do not group observations by view in the output.}
Therefore, the typical formulation of radiology report generation does not give the model sufficient information to generate its expected output.
This framework allows us 
to carefully consider what to include in the input and the output to develop reasonable problem formulations so that the model has sufficient information and that the evaluation focuses on the relevant components.

Following the pragmatic perspective, we provide a rigorous analysis to show that the indication drives negative mentions.
We thus reformulate the radiology report generation problem as
generating a radiology report given an image and an indication.
With respect to the output, we use large language models (LLMs) to clean the reports by removing uninferable information from the input,
which also redefines the desired generation output.
Accordingly, 
we introduce novel evaluation metrics 
to disentangle model limitation from uninferable information, negative mentions from correctness of positive findings.
Finally, we build pragmatic generation models and demonstrate substantial performance improvements compared to existing approaches. 
In particular, our LLaMA-based model, even when trained on unclean reports, produces fewer hallucinations than retrieval-based methods retrieving from cleaned data.

As a side outcome of our framework, our analysis reveals a clear distribution shift between the test set and the training set in 
MIMIC-CXR.
On average, each report has only 0.255 negative mentions in the test set, compared to 0.485 in the training set, which challenges the i.i.d assumption.
We recommend the community carefully rethink the use of the standard train-test split in MIMIC-CXR for benchmark purposes in the future.

In summary, we make the following contributions:
\begin{itemize}[itemsep=0pt,leftmargin=*]
    \item We introduce the pragmatic perspective and reformulate the problem of radiology report generation.
    \item We demonstrate that the indication drives mentions of negative observations and develop new evaluation metrics inspired by the pragmatic perspective.
    \item We show that our pragmatics-aware approaches lead to better generation, in both traditional and proposed evaluation metrics.
    \item We reveal idiosyncrasies in the test set of the MIMIC-CXR dataset.
\end{itemize}

Our code is available at \url{https://github.com/ChicagoHAI/llm\_radiology}.

\section{Dataset}
\label{sec:dataset}

We use MIMIC-CXR, a chest X-ray dataset containing 377,110 images and their corresponding reports~\citep{johnson2019mimic}. 
It has been widely used in recent studies on report generation~\citep{liu2019clinically, chen2020generating, miura2021improving,  endo2021retrieval, ramesh2022improving, thawkar2023xraygpt}, and comes with a train/dev/test split. 

Following prior work, we use CheXbert to derive groundtruth labels for each image based on the corresponding report~\citep{smit2020chexbert}.
For each report, there are fourteen conditions: twelve thoracic conditions, one condition for support devices, and one for No Finding. 
Except for No Finding, each condition can take four labels: 1 (positive), 0 (negative), -1 (uncertain), and missing (not mentioned). 
No Finding is either missing or 1.
Table~\ref{tab:neg_prop} presents basic statistics for the train/dev/test splits in MIMIC-CXR.

\begin{table}[t]
    \small
    \centering
    \begin{tabular}{@{}p{4.5cm}@{\hskip 8pt}r@{\hskip 8pt}r@{\hskip 8pt}r@{}}
    \toprule
    & Train & Dev & Test \\
    \midrule
    \#Reports & 371,951 & 1,837 & 2,872 \\
      \% No Finding & 41.3 & 40.9 & 19.8 \\
      avg. \#positive mentions & 1.35 & 1.17 & 1.39 \\
      avg. \#positive mentions in reports that are not ``No Finding'' & 1.59 & 1.29 & 1.49 \\
      avg. \#negative mentions & 0.485 & 0.232 & 0.255 \\
      avg. \#negative mentions in reports that are not ``No Finding'' & 0.826 & 0.394 & 0.318 \\
      \midrule
      \multicolumn{4}{c}{\% of reports that have negative mentions}\\
      \midrule
      Pneumothorax & 52.8 & 50.0 & 46.5 \\
      Pneumonia & 45.1 & 34.9 & 25.4 \\
      Edema & 44.3 & 25.5 & 20.2 \\
      Pleural Effusion & 45.4 & 19.4 & 26.1 \\
      Cardiomegaly & 42.9 & 29.6 & 27.4 \\
      Consolidation & 52.0 & 40.0 & 26.5 \\
      {\footnotesize Enlarged Cardiomediastinum} & 44.7 & 25.0 & 0.0 \\
      Lung Opacity & 50.7 & 26.7 & 16.0 \\
      Lung Lesion & 46.0 & 12.0 & 18.2 \\
      Fracture & 43.7 & 0.0 & 21.1 \\
      Support Devices & 50.3 & 17.0 & 31.5 \\
      Atelectasis & 43.5 & 15.4 & 15.0 \\
      Pleural Other & 42.9 & 0.0 & 0.0 \\
      No Finding & 34.0 & 15.1 & 16.2 \\
      \bottomrule
    \end{tabular}
    \caption{Top: Statistics on the positive and negative mentions of MIMIC-CXR. Bottom: Percentage of reports that contain at least one negative mention, conditioned on a condition mentioned in the indication. The conditions are sorted by the frequency of their negative mentions (see Appendix~\ref{sec:appendix_data}).
    }
    \label{tab:neg_prop}
\end{table}

\paragraph{Negative mentions are prevalent.}
On average, there is about one negative mention for every three positive mentions in the training set. 
When a report is not labeled ``No Finding'',  this ratio becomes less than one-to-two. 
This shows that commenting on negative observations is common practice in radiology reporting, a phenomenon that we will revisit in  \secref{sec:pragmatic}.

\paragraph{Substantial discrepancies between the training set and the test set.}
41.3\% of reports are ``No Finding'' in the training set, while only 19.8\% of the test set are ``No Finding''.
Furthermore, an average test report only contains 0.255 negative mentions compared to 0.485 in an average train report.
In contrast, the average numbers of positive mentions are similar between the training set and the test set.

We further group results by conditions identified in the indication\footnote{Positive, negative, and uncertain labels are all considered mentions.} with CheXbert in Table~\ref{tab:neg_prop}.
When a condition is mentioned in the indication, about half of the time the report has at least one negative mention 
in the training set, further confirming the importance of negative mentions.
Meanwhile, we observe a discrepancy between the training set and the test set: the percentage of negative mentions is much lower, often half of the rate as in the training set, with ``Pneumothorax'' as the only exception. 

This raises the question of whether the issue lies with the training or the test set. We briefly compared the same data statistics across two other datasets: CheXpert ~\citep{irvin2019chexpert} and OpenI ~\citep{demner2016preparing}, and found 
that their average numbers of negative mentions per report are much more similar to those of MIMIC-CXR's training set than its test set.
This gives evidence for the test set being out-of-distribution.
These results are further discussed in \secref{sec:conclusion}.
However, given that CheXpert does not have reports (although the labels for their images were derived from accompanying reports), and that OpenI is much smaller than MIMIC-CXR and does not have a test set, we decided to use MIMIC-CXR despite its discrepancies.

\section{Rethinking Radiology Report Generation Pragmatically}
\label{sec:pragmatic}

In this section, we start with a rigorous analysis of the connection between the indication and mentions of negative observations.
This analysis motivates our reformulation of the generation problem.
Then, building on our framework in Table~\ref{tab:input_factors}, we use large language models to clean reports to remove content that we do not expect models to generate given the image and the indication.
Finally, we introduce novel evaluation metrics inspired by this pragmatic perspective.

\subsection{A Pragmatic Observation of Indication and Negative Mentions}
\label{sec:neg_gen}

Consider a normal chest X-ray. 
Based on the image alone,
it is impossible to favor either of the following two reports: ``No acute cardiopulmonary process.'' and ``No radiographic evidence for pneumonia.''
Next, we show that the \emph{indication} section drives negative mentions like that in the second report.

Table~\ref{tab:neg_prop} has demonstrated the prevalence of negative mentions.
We would like to capture the probability of negative mentions given an indication instead of simply computing the percentage of negative mentions in the reports.
Leveraging the intuition from our example, the key idea is that the probability of negative mentions only makes sense in reports where the condition is actually not positive; in fact, a condition appearing in the indication increases the probability of the condition being positive, deflating the probability of negative mentions.
Therefore, we ignore these positive cases when computing the probability of negative mentions.

Specifically, for a report $R$, we denote its indication section as $I(R)$.
As discussed in \secref{sec:dataset}, for each condition $X$, the report is labeled as $R_X \in \{1, 0, -1, -2\}$,
where $1, 0, -1$ correspond to positive, negative, and uncertain mentions of the condition per CheXbert's convention, while $-2$ suggests the condition is not mentioned in $R$.
For every condition $X$ except No Finding, 
we compute two conditional probabilities depending on the event that $X$ appears in the indication, which is denoted $X \in I(R)$:
\begin{align*}
\mbox{\normalsize $P(\neg X \mid X \in I)  = \dfrac{|\{R: R_X = 0 \wedge X \in I(R) \}|}{|\{R: R_X \in \{0, -2\} \wedge X \in I(R) \}|}$},\\
\mbox{\normalsize $P(\neg X \mid X \notin I) = \dfrac{|\{R: R_X = 0 \wedge X \notin I(R) \}|}{|\{R: R_X \in \{0,-2\} \wedge X \notin I(R) \}|}$},
\end{align*}
where $\neg X$ refers to negative mentions of $X$, and $R \in \mathcal{R}$ the set of all reports.

\begin{table}[t]
    \centering
    \small
    \begin{tabular}{@{}p{3cm}@{}r@{}r@{}}

        \toprule
         Condition &  $P(\neg X \mid X \in I)$  & $P(\neg X \mid X \notin I)$ \\
        \midrule
        Atelectasis *** &  1.7\% & 0.3\% \\
        Cardiomegaly  &  6.2\% & 5.8\% \\
        Consolidation *** &  7.3\% & 3.3\% \\
        Edema *** &  23.4\% & 8.0\% \\
        \small{Enlarged Cardiomediastinum ***} & 8.6\% & 2.1\% \\
        Fracture *** & 14.0\% & 0.3\% \\
        Lung Lesion *** &  5.8\% & 0.4\% \\
        Lung Opacity *** & 2.2\% & 0.8\% \\
        Pleural Effusion *** &  18.1\% & 8.3\% \\
        Pleural Other *** &  0.9\% & 0.03\% \\
        Pneumonia *** &  25.0\% & 8.9\% \\
        Pneumothorax *** &  42.7\% & 9.1\% \\
        Support Devices *** & 3.7\% & 0.2\% \\
        \bottomrule
    \end{tabular}
    \caption{$\chi^2$-test results show that negative mentions are influenced by the indication. 
    *** indicates $p < 0.001$. No Finding is excluded.
    }
    \label{tab:p_vals}
\end{table} 

Table~\ref{tab:p_vals} shows the results and whether the differences between these two probabilities are significant based on the $\chi^2$-test on the training set. 
All differences are significant except for Cardiomegaly. 
For most conditions, $P(\neg X \mid X \in I)$ is substantially greater than $P(\neg X \mid X \notin I)$, which offers strong evidence that conditions are more likely to be mentioned as negative when they are inquired about in the indication.

Given the important role of indication in determining negative mentions, we reformulate the problem of radiology report generation as generating the report given an image and an indication.

\begin{table*}[ht]
\small
\centering
    \begin{tabular}{p{3.5cm}p{6.5cm}p{5.5cm}}
    \toprule
    Rule &  Original &  Cleaned \\
    \midrule
    Remove comparison to prior studies & In comparison with the study of \_\_\_, there are slightly improved lung volumes. & There are slightly improved lung volumes. \\
    \hline
    Remove communication information & These findings were communicated via the radiology critical results dashboard at 12:57 p.m. & REMOVED \\
    \hline
    Rewrite new/increased conditions into positive & New large right pneumothorax & Large right pneumothorax \\
    \hline
    Rewrite resolved conditions into negative & Resolved opacities in the left mid lung. & No opacities in the left mid lung. \\
    \bottomrule
  \end{tabular}
  \caption{Example cleaning rules. See Appendix~\ref{sec:appendix_cleaning} for details.}
  \label{tab:cleaning_rules_main}
\end{table*} \subsection{Pragmatic Data Cleaning}
\label{sec:preprocessing}

In addition to including indications as part of the input, we need to carefully consider what the desired output should include.
We focus on information that one can generate from the image and the indication in this work, so we aim to remove the following information in Table~\ref{tab:input_factors}: previous studies, previous treatment, recommendations,\footnote{We opt to be conservative in this work as this information often depends on the patient's preference and urgency.} doctor communications, image view.
Our framework is a generalization of previous attempts to clean reports~\citep{ramesh2022improving, thawkar2023xraygpt} which focus on removing references to prior studies and image views.

\paragraph{Methodology.}
We developed our method on a set of 100 manually cleaned reports.
Inspired by~\citet{thawkar2023xraygpt}, we use few-shot in-context learning to perform the cleaning.
Specifically, we create seven rules to remove the information of interest and prompt Flan-T5-XXL with a small number of examples
to clean reports~\citep{longpre2023flan} (see Table~\ref{tab:cleaning_rules_main} for examples).
We prompt the model using one rule at a time and refer to this approach as ``rule composition''. 
This approach provides more flexibility than the fine-tuned classifier (GILBERT) in~\citet{ramesh2022improving} and leverages the capability of LLMs to rewrite rather than remove information.
During the development of our method, we found that cleaning can change the CheXbert labels of a sentence, due to flaws in Flan-T5 and CheXbert, so we employ a simple heuristic after every cleaning step to discard the change if it has changed any label.

\paragraph{Evaluation.}
We manually cleaned another 160 sentences as a test set.
For evaluation, 
we compute Positive and Negative F1 (see Section~\ref{sec:metrics}) 
using the labels of the LLM-cleaned and original sentences to evaluate whether the cleaning process maintains the original labels.
We also compute Exact Match (EM) accuracy and BLEU-2 between LLM-cleaned and manually-cleaned sentences to evaluate the similarity at the token level.
In addition, we provide a heuristic measure for each type of uninferable information at the report level.
For an information type, we define a few keywords denoting a mention of that information. 
We calculate the percentage of reports that has such information after cleaning.
Details on the development and test sets that we use for Flan-T5 report cleaning can be found in Appendix~\ref{sec:measure_halluc}

\begin{table}[t]
\small
    \centering
    \resizebox{\linewidth}{!}{
    \begin{tabular}{@{}p{2.7cm}@{\hskip 8pt}>
    {\centering\arraybackslash}p{1cm}@{\hskip 4pt}>
    {\centering\arraybackslash}p{1.1cm}@{\hskip 4pt}>
    {\centering\arraybackslash}p{1.4cm}@{\hskip 4pt}>
    {\centering\arraybackslash}p{1.4cm}@{}}
    \toprule
     Model &  Pos F1 &  Neg F1 &  EM Acc. &  BLEU-2 \\
    \midrule
    GILBERT & 0.915 & 0.846 & 0.188 & 0.505 \\
    Flan-T5 (all-rules) & 0.930 & 0.898 & 0.419 & 0.514 \\
    Flan-T5  (compose-rules) & 0.855 & 0.821 & \bf 0.538 & 0.527 \\
    Flan-T5 (compose-rules + label heuristic) & \bf 1.000 & \bf 1.000 & 0.531 & \bf 0.541 \\ 
    \bottomrule
    \end{tabular}}
    \caption{Report cleaning result at the sentence level.}
    \label{tab:cleaning_results}
\end{table}

\tableref{tab:cleaning_results} reports our cleaning model's performance at the sentence level.
To test the effectiveness of rule composition, we compare our model against a Flan-T5 model prompted using all of the rules in one prompt.
All Flan-T5 variants outperform GILBERT, which is expected since the latter only cleans ``previous studies'' under our framework in Table~\ref{tab:input_factors}. 
Since Flan-T5 (compose-rules) outperforms Flan-T5 (all-rules),
rule composition is shown to be effective. 
With the label heuristic, we benefit from cleaning sentences without accidentally changing their meaning, despite the slightly lower accuracy.

\begin{table}[t]
\small
    \centering
    \resizebox{\linewidth}{!}{
    \begin{tabular}{l@{}>
    {\centering\arraybackslash}p{1.7cm}@{}>
    {\centering\arraybackslash}p{1.7cm}@{\hskip 8pt}>
    {\centering\arraybackslash}p{0.8cm}@{\hskip 8pt}>
    {\centering\arraybackslash}p{0.8cm}@{\hskip 8pt}>
    {\centering\arraybackslash}p{0.5cm}@{\hskip 8pt}}
    \toprule
    Model     & Prior study & Prior proc. & Comm. & Rec. & View \\
    \midrule
    Train     & 52.6\% & 1.2\% & 9.6\% & 10.5\% & 6.4\% \\
    GILBERT     & \bf 25.1\% & 1.1\% & 9.2\% & 10.3\% & 6.4\% \\
    XrayGPT     & 53.8\% & 1.5\% & 20.2\% & 22.2\% & 8.1\% \\
    Flan-T5     & 30.5\% & \bf 0.7\% & \bf 4.6\% & \bf 7.3\% & \bf 4.3\% \\
    \bottomrule
    \end{tabular}}
    \caption{Percentage of reports with uninferable information after cleaning. Lower is better.}
    \label{tab:train_hallu}
\end{table}
\begin{figure*}[ht]
    \centering
    \includegraphics[width=.75\textwidth]{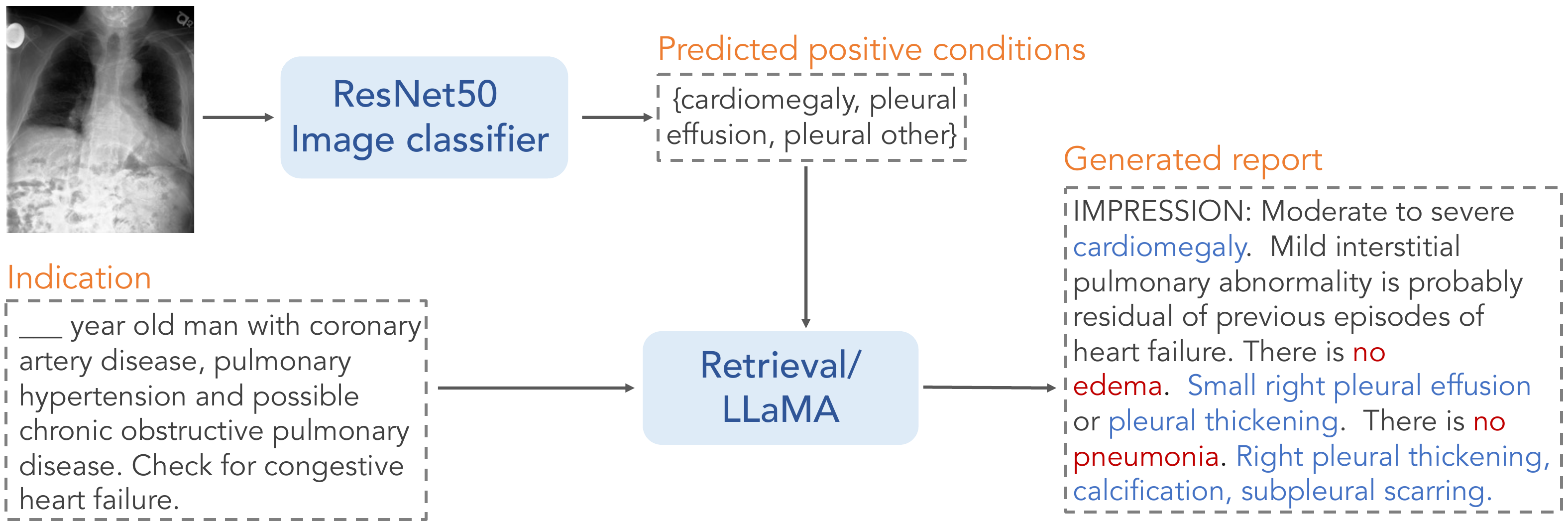}
    \caption{An overview of our approach. Blue: positive findings. Red: negative mentions.}
    \label{fig:model_diagram}
\end{figure*}

Table~\ref{tab:train_hallu} shows the extent to which Flan-T5 cleans uninferable information at the report level compared to other baselines.
It outperforms GILBERT and XrayGPT 
on cleaning all information types, with the only exception being prior studies, on which it trails behind GILBERT slightly.
Given the encouraging results, we employ Flan-T5 with rule composition and the label heuristic to clean MIMIC-CXR.

\subsection{Pragmatic Evaluation}
\label{sec:metrics}

We start by reviewing standard evaluation metrics.
\begin{itemize}[itemsep=0pt,leftmargin=*]
    \item Clinical efficacy (CE). We include Positive F1, Positive F1-5 that focuses on the most frequent five conditions,
    and RadGraph F1~\citep{jain2021radgraph}, as they are commonly used to evaluate the correctness of reports, and especially as RadGraph F1 has been shown to align well with radiologists' judgements~\citep{endo2021retrieval, irvin2019chexpert, yu2022evaluating}.
    \item Language performance against original reports. We use BLEU~\citep{papineni2002bleu} and BERTScore~\citep{zhang2019bertscore}, which are commonly used for natural language generation tasks. 
    As demonstrated in Table~\ref{tab:input_factors}, we do not think that these are appropriate metrics because much of the content is impossible to generate given the image.
    We keep these two metrics as they are standard in existing work.
\end{itemize}

Inspired by the pragmatic perspective, we believe that existing metrics are flawed in two ways:
1) comparing against original reports expects the model to generate uninferable information;
2) clinical efficacy ignores the prevalent mentions of negative observations.
Thus, we develop the following metrics to capture the pragmatic performance of report generation.
\begin{itemize}[itemsep=0pt,leftmargin=*]
    \item Clean BLEU-2 and Clean BERTScore. As some information is impossible to generate given the image and the indication, using the original report as the groundtruth is not ideal. We thus compute BLEU-2 and BERTScore against the cleaned reports.
    \item Negative F1 and Negative F1-5. 
    Parallel to Positive F1 and Positive F1-5, 
    we introduce Negative F1 and Negative F1-5, which evaluates against whether a negative mention occurs in the report for a particular label. For Negative F1-5, we use the most frequent five negative labels in the training set, as shown in Table~\ref{tab:neg_prop}: Pneumothorax, Pneumonia, Edema, Pleural Effusion, and Consolidation.\footnote{Although Cardiomegaly is more common than Consolidation in the training set, there are too few examples in the dev and test set so we exclude it. See Appendix~\ref{sec:appendix_data}.}
    \item Hallucination. 
    Finally, we measure how often the model generates information that cannot be generated from the image and the indication. For simplicity, we merge all uninferable information types into one measure with the keywords in \secref{sec:preprocessing} and compute the percentage of generated reports that contains any uninferable information.

\end{itemize}
Following \citet{endo2021retrieval}, all evaluations are done on the impression section.\footnote{Only 12.5\% of the reports have the Findings sections~\citep{johnson2019mimic}.}

\section{Experiments}
\label{sec:methods}

To demonstrate the practical importance of the pragmatic perspective, we perform experiments on radiology report generation.

\subsection{Method and Experiment Setup}

\paragraph{Our approach.}
Our approach disentangles predictions based on the image from generations based on the positive conditions (Figure~\ref{fig:model_diagram}).
We first use a ResNet-50~\citep{he2016resnet} to predict the positive conditions in the image. 
Then, leveraging the insight from \secref{sec:pragmatic}, we generate the reports based on the indication and the predicted positive conditions.
We consider two approaches in text generation:
\begin{itemize}[itemsep=0pt,leftmargin=*]
\item {\bf Pragmatic retrieval}. We first check the predicted labels against the conditions in the indication. For every condition in the indication that is not predicted as positive based on the image, we retrieve a cleaned sentence that mentions that condition
as negative. For the predicted positive conditions, we retrieve a report from the training set with the same set of positive conditions and concatenate it with the sentences with negative mentions to form the final report.
\item {\bf Pragmatic LLaMA}. We finetune LLaMA-7B~\citep{touvron2023llama} to generate clean reports using the predicted positive conditions and the indication as input.
We use the same hyperparameters as those of Alpaca~\citep{alpaca} and train on a sample of 18,264 unique, clean report impressions (10\% of the training data\footnote{We found that training on more data, e.g., 80\%, did not improve performance.}) for 3 epochs.
Our prompt can be found in Appendix~\ref{sec:appendix_model}.
\end{itemize}

\begin{table*}[t]
    \centering
    \small
    \begin{tabular}{@{}p{3cm}@{}>
    {\centering\arraybackslash}p{0.8cm}@{\hskip 4pt}>
    {\centering\arraybackslash}p{1cm}@{\hskip 4pt}>
    {\centering\arraybackslash}p{1.4cm}@{\hskip 4pt}|>
    {\centering\arraybackslash}p{1cm}@{\hskip 4pt}>
    {\centering\arraybackslash}p{1cm}@{\hskip 4pt}|>
    {\centering\arraybackslash}p{1.2cm}@{}>
    {\centering\arraybackslash}p{1.8cm}@{\hskip 4pt}>
    {\centering\arraybackslash}p{0.8cm}@{\hskip 8pt}>
    {\centering\arraybackslash}p{0.8cm}@{\hskip 4pt}>
    {\centering\arraybackslash}p{1.2cm}@{\hskip 4pt}}
    \toprule
    & \multicolumn{3}{c|}{\bf Correctness} & \multicolumn{2}{c|}{\bf Language} & \multicolumn{5}{c}{\bf Pragmatic metrics} \\
    Model &  Pos F1 &  Pos F1-5 &  Rad-Graph F1 &  BLEU-2 &  BERT-Score &  Clean BLEU-2 &  Clean BERTScore &  Neg F1 &  Neg F1-5 &  Halluci-nation$\downarrow$  \\
    \hline
    CXR-RePaiR & 0.238 & 0.368 & 0.076 & 0.027 & 0.162 & 0.028 & 0.176 & 0.016 & 0.042 & 0.756 \\
    CXR-ReDonE & 0.206 & 0.320 & 0.113 & 0.048 & 0.251 & 0.050 & 0.269 & 0.045 & 0.102 & 0.518 \\
    MedCLIP & 0.122 & 0.239 & 0.077 & 0.023 & 0.180 & 0.025 & 0.199 & 0.013 & 0.024 & 0.260 \\
    XrayGPT 0-shot & 0.074 & 0.056 & 0.013 & 0.007 & 0.005 & 0.007 & 0.012 & 0.014 & 0.028 & 0.578 \\
    \hline
    Pragmatic Retrieval & 0.293 & 0.403 & 0.103 & 0.072 & -0.103 & 0.078 & -0.084 & \bf 0.077 & \bf 0.156 & 0.445 \\
    Pragmatic LLaMA & \bf 0.307 & \bf 0.417 & \bf 0.194 & \bf 0.137 & \bf 0.360 & \bf 0.151 & \bf 0.385 & 0.050 & 0.127 & \bf 0.158 \\
    \bottomrule
    \end{tabular}
    \caption{Report generation performance. Pragmatic methods outperform all previous methods that do not make use of the indication section, in both traditional metrics and our pragmatics-inspired metrics.}
    \label{tab:test_results}
\end{table*}

\paragraph{Baselines.} We consider the following baselines from existing work. 
\begin{itemize}[itemsep=0pt,leftmargin=*]
    \item \textbf{CXR-RePaiR}~\citep{endo2021retrieval}: a model that retrieves $k=2$ report sentences that have the most similar embeddings to that of the image using cosine similarity. 
    The embeddings are learned using CLIP on MIMIC-CXR. 
    CXR-RePaiR is the state of the art on Positive F1 and Positive F1-5.\footnote{We omit a recent work from Google because we do not have access to their model~\citep{tu2023medpalmm}.}
    \item \textbf{CXR-ReDonE}~\citep{ramesh2022improving}: a model with a similar training and retrieval strategy as those of CXR-RePaiR, but is trained on CXR-PRO, a clean version of MIMIC-CXR by removing references to prior studies.
    CXR-ReDonE is the state of the art on RadGraph F1, BERTScore, and Hallucination.\footnote{Although MedCLIP hallucinates less, our manual inspection of its generated reports shows that it retrieves a small set of sentences for all test examples, which can trivially minimize hallucination, so we discount it.} 
    \item \textbf{MedCLIP}~\citep{wang2022medclip}: similar to CXR-RePaiR, but the image and text embeddings are learned using MedCLIP.
    \item \textbf{XrayGPT}~\citep{thawkar2023xraygpt}: a model consisting of a vision encoder and a LLM decoder. 
    The representations between the two modalities are aligned using a fully-connected (FC) layer in-between the encoder and decoder. 
    The FC layer is trained using MIMIC-CXR and OpenI data~\citep{demner2016preparing}. 
    XrayGPT also cleans prior studies using gpt-3.5-turbo and rules via prompting~\citep{thawkar2023xraygpt}.
\end{itemize}

\paragraph{Evaluation.} We use both standard and  pragmatics-inspired metrics defined in \S\ref{sec:metrics}.

\subsection{Performance Comparisons}
\label{sec:results}

\paragraph{Pragmatic models outperform previous non-pragmatic methods on all metrics (Table~\ref{tab:test_results}).}
Pragmatic-LLaMA outperforms all the baselines by a substantial margin in both traditional and pragmatic metrics.
On Positive F1, our model outperforms CXR-RePaiR by 6.9\% in absolute score and 29\% relatively. 
It also surpasses CXR-ReDonE at RadGraph F1 and BERTScore by 8.1\% points (+71\% relative) and 0.109 points (+43\% relative), respectively. 
On Hallucination, only 15.8\% of Pragmatic-LLaMA's reports contain hallucinations, a 69.5\% reduction from CXR-ReDonE.
With respect to Negative F1-5, Pragmatic Retrieval is the best model, surpassing Pragmatic-LLaMA by 2.9\% points.
Overall, both of our pragmatic models outperforming non-pragmatic methods in all metrics demonstrates the effectiveness of using the indication as input, not only for negative mention generation, but also for clinical efficacy and mimicking radiologist writing style.

Although Pragmatic-LLaMA trails behind Pragmatic Retrieval in negative mention metrics,
it still outperforms the latter by a large margin on metrics that assess language similarity to the groundtruth report, such as RadGraph F1, BLEU-2, BERTScore, and Hallucination.
We believe that the slight improvement of Pragmatic Retrieval over Pragmatic-LLaMA in Negative F1 is because the negative mention distribution of the test set is vastly different from that of the training set, as shown in Table~\ref{tab:neg_prop}.
Thus, this number may not accurately reflect Pragmatic-LLaMA's negative mention generation performance. 
We thus evaluate the models on reports with Pneumothorax in the indication, the label with the least discrepancy between the negative proportion in the training and the test set.
Indeed, when the test set is in-distribution with respect to the training set, Pragmatic-LLaMA outperforms Pragmatic Retrieval by 7.7 points on Negative F1-5 (+40.3\% relative). 
At the same time, it remains the best model at almost every other metric (see Appendix~\ref{sec:appendix_results}).

\subsection{Ablation Results}
\label{sec:ablations}

We conduct ablation experiments to identify the effect of 1) incorporating the indication in the input and 2) report cleaning.
We denote Pragmatic-LLaMA's modifications by ``Cleaning + Indication''. 
We compare it with ``Cleaning Only'' 
and ``Indication Only''.
All other variables in training these models are controlled.
Table~\ref{tab:ablation} shows the results.

\begin{table}[t]
    \small
    \begin{tabular}{p{3.01cm}>{\centering\arraybackslash}p{0.7cm}>{\centering\arraybackslash}p{0.7cm}>{\centering\arraybackslash}p{0.7cm}>{\centering\arraybackslash}p{0.8cm}}
        \toprule
         Model & Pos F1-5 & BERT-Score & Neg F1-5 & Halluci-nation$\downarrow$ \\
         \midrule
         Cleaning+Indication & 0.427 & 0.479 & 0.099 & 0.185 \\
         Indication Only & 0.404 & 0.464 & 0.096 & 0.322 \\
         Cleaning Only & 0.417 & 0.464 & 0.065 & 0.128 \\
         \bottomrule
    \end{tabular}
    \caption{Ablation study. We select these four representative metrics for space reasons. The full table of results can be found in Appendix~\ref{sec:appendix_ablation}.}
    \label{tab:ablation}
\end{table}

First, 
comparing Cleaning + Indication with Indication Only shows that 
not only does cleaning help reduce the number of hallucinations in the output, but it also improves model performance on BERTScore and Positive F1-5. 
It could be because cleaning helps remove noise from the training data, which simplifies the learning task and helps the model generate cleaner outputs, in turn allowing CheXbert to label the generated reports more correctly.  

Second, Cleaning Only achieves lower scores than Cleaning + Indication, suggesting that adding the indication improves model performance, especially on Negative F1.
This further shows that the indication can help a model generate negative mentions. 
We did not expect the model to improve on other metrics, as they are limited by the vision model's ability. 
But from our inspection of the data set, we found that sometimes the impression will repeat a few words from the indication.
Thus, when a model is trained with the indication in the input, it learns to repeat words from the indication, leading to an increase in BERTScore. 
As for CE metrics, since LLaMA is imperfect, it may fail to report conditions even though they are explicitly given in the prompt. 
In these cases, the indication can provide extra signal to ``remind'' the model to include the prompted conditions.

Interestingly, adding the indication increases hallucinations, as that proportion is higher for Cleaning + Indication compared to Cleaning Only. 
It is likely because some indications refer to previous studies and procedures as context for the current study. 
The model then refers to this information when it generates the impression. 
We provide evidence for this claim and discuss it more in Appendix~\ref{sec:appendix_ablation} with a breakdown of the types of hallucination generated and some examples.
In short, adding the indication does not make the model generate more recommendations, but it makes the model generate more comparisons, showing that it likely repeats information from the indication.
Even so, Indication Only, our fine-tuned model with the most hallucinations, still produces fewer hallucinations than CXR-ReDonE, which retrieves from cleaned data.
That is, despite the common perception that language models like LLaMA are prone to hallucinations when generating radiology reports~\citep{ji2023hallucination}, LLaMA is 
more resistant to hallucinations compared to retrieval-based methods.

 \begin{table*}[!ht]
    \centering
    \small
    \begin{tabular}{@{}p{9cm}@{\hskip 8pt}r@{\hskip 8pt}r@{\hskip 8pt}r@{\hskip 8pt}r@{\hskip 8pt}r@{\hskip 8pt}r@{}}
    \toprule
    & \multicolumn{3}{c}{\bf MIMIC-CXR} & \multicolumn{2}{c}{\bf CheXpert} & \bf OpenI \\
    & Train & Dev & Test & Train & Test & Train \\
    \midrule
    \#Reports & 371,951 & 1,837 & 2,872 & 223,414 & 500 & 3,955 \\
      \% No Finding & 41.3 & 40.9 & 19.8 & 0.0 & 0.0 & 59.7 \\
      avg. \#positive mentions & 1.35 & 1.17 & 1.39 & 2.189 & 1.894 & 0.970 \\
      avg. \#positive mentions in reports that are not ``No Finding'' & 1.59 & 1.29 & 1.49 & 2.190 & 1.894 & 0.925 \\
      avg. \#negative mentions & 0.485 & 0.232 & 0.255 & 0.918 & 1.166 & 0.330 \\
      avg. \#negative mentions in reports that are not ``No Finding'' & 0.826 & 0.394 & 0.318 & 0.918 & 1.166 & 0.819 \\
      \bottomrule
    \end{tabular}
    \caption{Dataset statistics related to positive and negative mentions across different datasets.
    }
    \label{tab:other_datasets}
\end{table*}

\section{Related Work}
\label{sec:related}

We briefly review related work from the pragmatic perspective. 
Most previous methods have framed the problem as captioning a single image, and focused on evaluating the correctness of positive observations.
Vision encoder-language decoder architectures have been shown to generate stylistically accurate reports, but with limited positive mention correctness~\citep{jing2018automatic, chen2020generating, boag2020baselines}.
In contrast, retrieval-based models sacrifice some coherence in favor of clinical efficacy~\citep{endo2021retrieval, ramesh2022improving}.
Going beyond single-image captioning, some works have attempted to model \emph{multiple} image views~\citep{yuan2019multiview, miura2021improving, lee2023unixgen}, which can potentially learn the image view information.
Regarding comparisons to prior studies, \cite{ramesh2022improving} and \cite{thawkar2023xraygpt} notice that such information in groundtruth reports can lead models to hallucinate about non-existent studies, and opt to remove them from the output.
To our knowledge, we are the first work to introduce a unified pragmatic framework and emphasize negative mentions.

\section{Concluding Discussion}
\label{sec:conclusion}
In this work, we introduce a new, pragmatic perspective on the problem of radiology report generation. 
We found that radiology reports contain important information beyond positive observations, and focused on generating negative mentions as a first step towards 
pragmatic report generation.
We show that the indication section is critical to 
reporting negative conditions, and by incorporating it in our models' input,
we outperform existing approaches on Negative F1 scores, Hallucination, as well as other standard metrics.
We encourage future work to take the new problem formulation and advance modeling approaches to further improve report generation and reduce hallucination.

Following the pragmatic perspective, we found that MIMIC-CXR may not be entirely suitable for training and evaluating models in radiology report generation. 
Table~\ref{tab:other_datasets} shows the dataset statistics on positive and negative mentions for two other datasets than MIMIC-CXR: CheXpert and OpenI. 
CheXpert only exposes chest \xray images to the user, while OpenI does not have a development or test split. 
Moreover, CheXpert deliberately limits reports without any finding, while OpenI probably samples the data more similarly to MIMIC-CXR. 
Due to this peculiarity of CheXpert, we now only refer to reports that are not ``No Finding'' when discussing these statistics. 
The average number of positive mentions are somewhat similar between the three datasets, with one to two positive observations per report. 
In contrast, CheXpert and OpenI have about one negative observation per report, which suggests they are much more similar to MIMIC-CXR's training set than its development or test set. 
As mentioned above, we believe this is evidence for the development and test set being out-of-distribution,
which renders MIMIC-CXR an {\em inappropriate} benchmark for evaluating the quality of radiology report generation.

We also recognize that our Hallucination metric is a simple heuristic for measuring a complex and major concern regarding the use of language models in healthcare. 
It likely underestimates the percentage of reports that contain hallucinations.
We believe that, going forward, viable reports must not only include the correct observations, but also limit hallucinations. 
Hence, much more work is needed in developing better metrics for hallucination in generated reports. 
Since the emphasis of our work is to introduce the pragmatics perspective, we opted to use the simple Hallucination heuristic for model comparison, and leave as future work the development of a more accurate and more clinically relevant metric.

\section*{Acknowledgments}

We would like to thank Lydia Chelala for her helpful insights about the radiology report writing process, Brent DeVries for his early work on the project, Chenghao Yang, Colin Hudler, and David Reber for technical assistance, Mourad Heddaya, Jiamin Yang, and members of the Chicago Human+AI Lab who have given us valuable input and feedback.
This paper is supported by in part by
a CDAC discovery grant at the University of Chicago and NSF grants IIS-2040989 and IIS-2126602.

\bibliography{nguyen23}
\newpage

\appendix

\section{Per-label Negative Mention Frequencies}
\label{sec:appendix_data}

\begin{table*}[htbp]
    \centering
    \begin{tabular}{lrrrrrr}
    \toprule
    & \multicolumn{3}{c|}{\bf Negative Mentions} & \multicolumn{3}{c}{\bf Indication Mentions} \\
    Condition & Train & Dev & Test & Train & Dev & Test\\
    \midrule
      Pneumothorax & 37,840 & 69 & 124 & 19,971 & 50 & 87 \\
      Pneumonia & 37,635 & 152 & 235 & 62,881 & 256 & 435 \\
      Edema & 30,110 & 62 & 133 & 36,999 & 159 & 228 \\
      Pleural Effusion & 26,667 & 25 & 61 & 27,117 & 88 & 163 \\
      Cardiomegaly & 18,794 & 14 & 9 & 17,511 & 77 & 97 \\
      Consolidation & 12,614 & 66 & 132 & 20,933 & 123 & 111 \\
      {Enlarged Cardiomediastinum} & 7,716 & 11 & 11 & 2,946 & 16 & 5 \\
      Lung Opacity & 2,844 & 4 & 8 & 20,586 & 100 & 111 \\
      Lung Lesion & 1,991 & 7 & 3 & 11,543 & 42 & 44 \\
      Fracture & 1,925 & 17 & 6 & 9,545 & 13 & 25 \\
      Support Devices & 1,213 & 0 & 9 & 35,036 & 107 & 149 \\
      Atelectasis & 938 & 0 & 2 & 6,493 & 17 & 31 \\
      Pleural Other & 107 & 0 & 0 & 583 & 0 & 2 \\
      No Finding & 0 & 0 & 0 & 70,489 & 709 & 1368 \\
      \bottomrule
    \end{tabular}
    \caption{Per-label negative mention frequencies in MIMIC-CXR's Train-Dev-Test sets.}
    \label{tab:neg_freq}
\end{table*}

In Table~\ref{tab:neg_freq}, the training set has six conditions that have significantly more negative mentions than others. They are Pneumothorax, Pneumonia, Edema, Pleural Effusion, Cardiomegaly, and Consolidation. This is reflected somewhat in the dev and test set, except Cardiomegaly, which takes up a much smaller portion of those sets compared to in the train set. Because of this reason, as mentioned in Section~\ref{sec:metrics}, we exclude Cardiomegaly from the Negative F1-5 metric.

\section{Cleaning Details}
\label{sec:appendix_cleaning}

Table~\ref{tab:cleaning_rules} shows our seven cleaning rules and examples of sentences before and after cleaning, and Table~\ref{tab:cleaning_prompts} shows the prompts that we use for each rule for cleaning. Some information are easy to clean, while others are harder. For instance, communication and recommendations often span an entire sentence, so we already achieve good results by removing the entire sentence. However, as evident from the prompt of rule 3, we found that Flan-T5 sometimes have trouble understanding which sentence constitutes a recommendation, so we provided it with a simple heuristic to, at the very least, remove sentences the contain the string ``recommend''.

We found that the most difficult information to remove is comparisons to prior studies, because it requires a nuanced understanding of time and how conditions change. On the one hand, there are explicit cues, such as when the radiologist prefaces a finding by saying that he/she is making an observation in comparison with a specific previous study, e.g., ``Compared to a previous study on [insert date], [insert finding]''. In this case, we employ rule 1 to remove that phrase.

However, what comes after the preface is much more challenging. The overall idea is we want to rewrite any mention of condition progression into present tense: the \xray either shows or does not show that condition. The first type of progression to consider is when a condition is new or worsened, in which case it should only be reported as present or positive. We use rule 5 to handle that case. The second type is when conditions improve but have not disappeared completely, which means they are still present. This is handled by rule 6. Lastly, when a condition is completely resolved, as it is not present in the \xray, it should be reported as negative. This is rule 7, which the model struggles with greatly, as it has to rewrite the sentence the most. In the examples of rule 5 and 6, although the rule itself requires a nuanced understanding, in practice, to apply it, often the model only has to remove parts of the sentence. In contrast, in the example of rule 7, it not only has to remove the word ``resolved'', but it also has to replace it with the word ``no''. Nevertheless, like with recommendations, it is non-trivial for a language model to understand what constitutes a condition progression, so we also supplied it with certain keywords to help give it signal on which sentence should be modified.

Another nuanced issue is how to apply rule 5 and 6 when the change refers to an organ instead of condition. In contrast to conditions, organs are always ``positive''. When a radiologist reports ``The heart has increased since \_\_\_'', it would be strange to rewrite it into ``The heart'' or ``The heart is positive'' according to a naive application of rule 5. That is why we opted to keep all mentions of changes to organs the same.

The final issue is we use a single rule 4 to clean both \xray view and prior procedures. In fact, mentions of prior procedures are probably the most difficult pragmatic information for a model to identify, because of the varied semantics of what a procedure is. There is no easy keyword heuristic to rely on either, since ``prior'' part is often implicit. For example, such a sentence could look like ``The patient has received a tube to remove air from their pleural space, and now there is no pneumothorax.'' There is no easy heuristic to identify the first clause, and in our experience, the model struggles greatly with understanding what constitutes a medical procedure. We only found one phrase that radiologists often use to talk about the state of the patient after a procedure: ``status post''. An example similar to the one above is: ``The patient is status post ET tube removal. No pneumothorax.'' Since the model has a low success rate on this rule, and there is only one viable keyword heuristic, we opted to combine it with the removal of image view---another simple rule that does not warrant its own prompt---into one prompt to save compute time, as processing the entire MIMIC-CXR dataset using a large language model is very time-consuming even just with a single rule.

\begin{table*}[htbp]
\setlength{\extrarowheight}{3pt}
\small
\resizebox{\textwidth}{!}{
    \begin{tabular}{p{.3cm}p{3cm}p{6cm}p{5.5cm}}
    \hline
    \bf ID & \bf Rule & \bf Original & \bf Cleaned \\
    \hline
    1 & Remove comparison to prior studies & In comparison with the study of, there are slightly improved lung volumes. & There are slightly improved lung volumes. \\
    \hline
    2 & Remove communication information & These findings were communicated via the radiology critical results dashboard at 12:57 p.m. & REMOVED \\
    \hline
    3 & Remove doctor recommendations & Recommend advising patient to avoid palpating the area to avoid irritating it. & REMOVED \\
    \hline
    4 & Remove previous treatment and image view & Small lateral pneumothorax is present in this patient status post right first rib resection. & Small lateral pneumothorax is present in this patient \\ 
    && Lateral view raises concern for pneumonia at the left lung base & Concern for pneumonia at the left lung base \\
    \hline
    5 & Rewrite new/increased conditions into positive & New large right pneumothorax & Large right pneumothorax \\
    && Mild interval increase in loculated right pleural effusion & Loculated right pleural effusion. \\
    \hline
    6 & Rewrite unchanged/partially-improved conditions into positive & Small right pleural effusion probably unchanged since & Small right pleural effusion \\
    && Mild pulmonary edema appears slightly improved & Mild pulmonary edema \\
    \hline
    7 & Rewrite resolved conditions into negative & Resolved opacities in the left mid lung. & No opacities in the left mid lung. \\
    \hline
  \end{tabular}}
  \caption{Cleaning rules and examples.}
  \label{tab:cleaning_rules}
\end{table*} 
\begin{table*}[htbp]
    \centering
    \small
    \begin{tabular}{p{1.0\textwidth}}
        \toprule
        Rule 1: \\
        You will be given a sentence from a chest X-ray report. Remove ALL sentences that contain comparisons to the past, and rewrite sentences minimally to preserve meaning. If a sentence contains the word "compare", remove it. If a sentence is empty after cleaning, replace it with the token "REMOVED". If a sentence contains "REMOVED", do not change it. \\
        \\
        Rule 2: \\
        You will be given a sentence from a chest X-ray report. Remove ALL sentences that contain information about communication between medical professionals, such as between doctors or nurses. If a sentence is empty after cleaning, replace it with the token "REMOVED". If a sentence contains "REMOVED", do not change it. \\
        \\
        Rule 3: \\
        You will be given a sentence from a chest X-ray report. Remove ALL sentences that mention medical recommendations from doctors. Remove sentences that contain "recommend". If a sentence is empty after cleaning, replace it with the token "REMOVED". If a sentence contains "REMOVED", do not change it. \\
        \\
        Rule 4: \\
        You will be given a sentence from a chest X-ray report. Remove ALL sentences that mention the chest X-ray view (e.g. AP, PA, lateral) or "status post". Rewrite sentences minimally to preserve meaning. If a sentence is empty after cleaning, replace it with the token "REMOVED". If a sentence is empty or contains "REMOVED", do not change it. \\
        \\
        Rule 5: \\
        You will be given a sentence from a chest X-ray report. Remove all instances of "new", "increase", "greater", "worsen", etc. and rewrite the sentence to preserve meaning. If the sentence mentions changes to an organ (e.g. lung, heart), do not rewrite it. If a sentence contains "REMOVED", do not change it. \\
        \\
        Rule 6: \\
        You will be given a sentence from a chest X-ray report. If a sentence mentions that a positive medical condition is unchanged or improved (but still positive), remove words related to "unchanged" or "improve" and rewrite the sentence to only say the condition. Otherwise, keep it the same. If a sentence contains "REMOVED", do not change it. \\
        \\
        Rule 7: \\
        You will be given a sentence from a chest X-ray report. If the sentence mentions the resolution or disappearance of a condition, rewrite it to simply say the condition is negative. Otherwise, keep the sentence the same. If a sentence is empty or contains "REMOVED", do not change it. \\
        \\
        \{EXAMPLES\} \\
        \\
        Original: \\
        \{INPUT\_QUERY\} \\
        New: \\
        \\
        \bottomrule
    \end{tabular}
    \caption{Prompts for report cleaning. In implementation, the few-shot examples and input query are inserted after every prompt. See Table~\ref{tab:cleaning_rules} for examples of how sentences are cleaned.}
    \label{tab:cleaning_prompts}
\end{table*}

\section{Flan-T5 Development and Testing}
\label{sec:measure_halluc}

The development set consists of 100 report sentences, with 20 in each major information category: prior comparisons, recommendations, communication, view and previous procedures, and no change. Our decision to group view and previous procedures is explained in Appendix~\ref{sec:appendix_cleaning}. For the test set, we procure 160 report sentences from eight categories: seven categories according to the rules in Table~\ref{tab:cleaning_rules}, and the eighth category of unchanged sentences. Similar to the development set, each category contains 20 sentences.

Table~\ref{tab:hallu_keywords} shows the keywords used to compute the percentage of reports containing hallucinations of each type. Specifically, for each type of uninferable information, a report contains it if the report contains any of its corresponding keywords. For prior comparisons, we developed our own keywords, as well as using those identified by~\cite{ramesh2022improving}. Keywords from other types are introduced by us. As mentioned in Appendix~\ref{sec:appendix_cleaning}, prior procedures are the most difficult information to identify, and the only consistent common keyword we found was ``status'' for ``status post''.

\begin{table*}[htbp]
    \centering
    \begin{tabular}{lc}
        \toprule
        Information Type & Keywords \\
        \midrule
        Prior Comparisons & compar, interval, new, increas, worse, chang, \\
        & persist, improv, resol, disappear, prior, stable, previous, again, \\ & remain, remov, similar, earlier, decreas, recurr, redemonstrate \\
        Prior procedures & status \\
        Communication & findings, commun, report, convey, relay, enter, submit \\
        Image view & ap, pa, lateral, view \\
        Recommendations & recommend, suggest, should \\
        \bottomrule
    \end{tabular}
    \caption{Heuristic keywords used to identify hallucinations in reports}
    \label{tab:hallu_keywords}
\end{table*}

\section{Model Details}
\label{sec:appendix_model}

Table~\ref{tab:llama_prompt} describes the prompt that we use to train and perform inference with our Pragmatic-LLaMA model. The first two sentences are kept the same from the prompt provided by~\cite{alpaca}. We keep the task interpretation open and only ask the model to respond to the indication instead of asking it to only generate negative mentions based on the indication. This likely explains the phenomenon where the model echoes contextual information from the indication, helping it achieve higher Positive F1 and BERTScore as mentioned in Section~\ref{sec:ablations}.

Another advantage of our Pragmatic-LLaMA model is interpretability, since we decouple the vision and language component. Our use of predicted vision labels to prompt the language model can be seen as using sparse image representations as opposed to dense ones in end-to-end models. This makes it easier to interpret the positive mentions generated by the language model, which is an important quality for clinical models.

Interestingly, using predicted labels as image representation for retrieval helps the Pragmatic Retrieval model achieve higher clinical efficacy than dense representation methods like CXR-RePaiR or MedCLIP. However, in theory, we believe retrieval with dense representations is still more expressive than with sparse representations, since finer-grained information, such as condition severity and location, can be matched between the image and sentence. This applies to generative models like Pragmatic-LLaMA as well, and we leave this investigation for future work.

\begin{table*}[htbp]
    \centering
    \begin{tabular}{p{0.8\textwidth}}
        \toprule
        Below is an instruction that describes a task, paired with an input that provides further context. \\
        Write a response that appropriately completes the request. \\
        \\
        \#\#\# Instruction: \\
        Write a radiology report responding to the indication. Include all given positive labels. \\
        \\
        \#\#\# Input: \\
        Indication: [insert indication]\\
        Positive labels: [insert positive labels in English]\\
        \\
        \#\#\# Response: \\
        \\
        \bottomrule
    \end{tabular}
    \caption{Prompt used for Pragmatic-LLaMA training and inference.}
    \label{tab:llama_prompt}
\end{table*}

\section{Full Ablation Results}
\label{sec:appendix_ablation}

\begin{table*}[t]
    \centering
    \small
    \begin{tabular}{p{3.5cm}@{}>
    {\raggedleft\arraybackslash}p{0.8cm}@{\hskip 4pt}>
    {\raggedleft\arraybackslash}p{0.8cm}@{\hskip 4pt}>
    {\raggedleft\arraybackslash}p{0.8cm}|>
    {\raggedleft\arraybackslash}p{0.8cm}>
    {\raggedleft\arraybackslash}p{0.8cm}|>
    {\raggedleft\arraybackslash}p{0.8cm}@{\hskip 4pt}>
    {\raggedleft\arraybackslash}p{0.8cm}|>
    {\raggedleft\arraybackslash}p{0.8cm}@{\hskip 4pt}>
    {\raggedleft\arraybackslash}p{1cm}>
    {\raggedleft\arraybackslash}p{0.8cm}>
    {\raggedleft\arraybackslash}p{0.8cm}>
    {\raggedleft\arraybackslash}p{0.8cm}
    }
    \toprule
    & \multicolumn{3}{c|}{\bf Correctness} & \multicolumn{2}{c|}{\bf Language} & \multicolumn{2}{c|}{\bf Pragmatics} & \multicolumn{5}{c}{\bf Hallucination (\%)} \\
    Model &  Pos F1 &  Pos F1-5 &  RG F1 &  BL-2 &  BERT-Score &  Neg F1 &  Neg F1-5 &  Prior study & Prior Proc. & Com. & Rec. & View  \\
    \hline
    Cleaning + Indication & 0.318 & 0.427 & 0.32 & 0.274 & 0.479 & 0.046 & 0.099 & 13.010 & 0.109 & 0.490 & 1.742 & 1.415 \\
    Indication Only & 0.319 & 0.404 & 0.319 & 0.267 & 0.464 & 0.071 & 0.096 & 23.462 & 0.054 & 0.544 & 3.484 & 1.960  \\
    Cleaning Only & 0.312 & 0.417 & 0.299 & 0.259 & 0.464 & 0.025 & 0.065 & 7.240 & 0.000 & 0.000 & 2.177 & 1.143 \\
    \bottomrule
    \end{tabular}
    \caption{Pragmatic-LLaMA ablation results.}
    \label{tab:ablation_full}
\end{table*} 
\begin{table*}
    \centering
    \begin{tabular}{p{8cm}p{8cm}}
        \toprule
         Indication & Generated report \\
         \midrule
         \_\_\_-year-old female patient with adenocarcinoma, status post left pleural effusion drain on \_\_\_. Study requested for evaluation of new pleural effusion. & Left-sided pleural effusion has \emph{decreased} in size… \\
         & \\
         Hypoxia and respiratory distress, evaluate for interval changes and consolidation vs. pleural effusion. & \emph{Interval worsening} of multifocal opacities, right greater than left… \\
         & \\
         \_\_\_ year old man with significant hypoxa, ?PCP PN\_\_\_. Evaluate interval change. & No significant \emph{change. Persistent} right upper lobe opacity. \\
         & \\
         \_\_\_ year old man with previous pneumothorax; pigtail catheter pulled yesterday // ?pneumothorax & \emph{Interval removal} of the left pigtail chest tube without evidence of pneumothorax… \\
         \bottomrule
    \end{tabular}
    \caption{Some examples of hallucinations arising from the indication section. Italics denote keywords that contribute to the sentence being classified as a hallucination.}
    \label{tab:ind_hallucination}
\end{table*} 
Table~\ref{tab:ablation_full} shows the ablation results for Pragmatic-LLaMA. We observe that adding the indication improves negative mention generation and cleaning helps reduce hallucination. While it seems like adding the indication increases hallucination, a breakdown of the types of ``hallucination'' generated shows that Pragmatic-LLaMA does not generate more recommendations, but it does so for every other types of pragmatic information. When inspecting model-generated reports, we found that sometimes the indication would include results from previous studies, procedures, previous information transmission, and the imaging technique for the current study to provide context for the report. During finetuning, LLaMA learns to copy this information directly into a report, which explains why those types of information are more prevalent while recommendations are not. We provide some examples of this phenomenon in table~\ref{tab:ind_hallucination}. For instance, in the last example, the model mentions the interval removal of the pigtail chest tube likely because of the phrase "pigtail catheter pulled yesterday" in the indication.

\section{Test Performance on Pneumothorax Examples}
\label{sec:appendix_results}

\begin{table*}[htb]
    \centering
    \small
    \begin{tabular}{p{2.9cm}@{}>
    {\centering\arraybackslash}p{1cm}@{\hskip 4pt}>
    {\centering\arraybackslash}p{1cm}@{\hskip 4pt}>
    {\centering\arraybackslash}p{1cm}|>
    {\centering\arraybackslash}p{1cm}>
    {\centering\arraybackslash}p{1cm}|>
    {\centering\arraybackslash}p{1cm}@{\hskip 4pt}>
    {\centering\arraybackslash}p{1cm}>
    {\centering\arraybackslash}p{1cm}>
    {\centering\arraybackslash}p{1cm}>
    {\centering\arraybackslash}p{1cm}}
    \toprule
    & \multicolumn{3}{c|}{\bf Correctness} & \multicolumn{2}{c|}{\bf Language metrics} & \multicolumn{5}{c}{\bf Pragmatic metrics} \\
    Model &  Pos F1 &  Pos F1-5 &  RG F1 &  BL-2 &  BScore &  Clean BL-2 &  Clean BScore &  Neg F1 &  Neg F1-5 &  Hallu-cination  \\
    \hline
    CXR-RePaiR, k=2 & 0.205 & \bf 0.387 & 0.093 & 0.037 & 0.144 & 0.036 & 0.167 & 0.004 & 0.010 & 0.901 \\
    CXR-ReDonE, k=2 & 0.222 & 0.316 & 0.097 & 0.050 & 0.241 & 0.059 & 0.270 & 0.025 & 0.066 & 0.605 \\
    MedCLIP & 0.133 & 0.264 & 0.053 & 0.012 & 0.139 & 0.011 & 0.160 & 0.0 & 0.0 & 0.111 \\
    XrayGPT 0-shot & 0.057 & 0.060 & 0.011 & 0.004 & -0.007 & 0.006 & 0.005 & 0.010 & 0.027 & 0.610 \\
    \hline
    Pragmatic retrieval & 0.272 & 0.392 & 0.068 & 0.054 & 0.129 & 0.066 & 0.167 & 0.074 & 0.191 & 0.655 \\
    Pragmatic-LLaMA & \bf 0.301 & 0.377 & \bf 0.168 & \bf 0.103 & \bf 0.327 & \bf 0.128 & \bf 0.370 & \bf 0.103 & \bf 0.268 & \bf 0.287 \\
    \bottomrule
    \end{tabular}
    \caption{Pragmatic-LLaMA results on the test set of reports with Pneumothorax in the indication compared with baselines.}
    \label{tab:test-ptx_results}
\end{table*} 
Table~\ref{tab:test-ptx_results} shows the performance on the subset of reports with indications mentioning ``pneumothorax''.
When the test set is in-distribution with the training set, Pragmatic-LLaMA outperforms Pragmatic Retrieval by 7.7 points on Negative F1-5 (+40.3\% relative). At the same time, it remains the best model at almost every other metric. 
\end{document}